# StackOverflow vs Kaggle: A Study of Developer Discussions About Data Science

David Hin
University of Adelaide
a1720858@adelaide.edu.au

**ABSTRACT**
Software developers are increasingly required to understand fundamental Data science (DS) concepts. Recently, the presence of machine learning (ML) and deep learning (DL) has dramatically increased in the development of user applications, whether they are leveraged through frameworks or implemented from scratch. These topics attract much discussion on online platforms. This paper conducts large-scale qualitative and quantitative experiments to study the characteristics of 197836 posts from StackOverflow and Kaggle. Latent Dirichlet Allocation topic modelling is used to extract twenty-four DS discussion topics. The main findings include that TensorFlow-related topics were most prevalent in StackOverflow, while meta discussion topics were the prevalent ones on Kaggle. StackOverflow tends to include lower-level troubleshooting, while Kaggle focuses on practicality and optimising leaderboard performance. In addition, across both communities, DS discussion is increasing at a dramatic rate. While TensorFlow discussion on StackOverflow is slowing, interest in Keras is rising. Finally, ensemble algorithms are the most mentioned ML/DL algorithms in Kaggle but are rarely discussed on StackOverflow. These findings can help educators and researchers to more effectively tailor and prioritise efforts in researching and communicating DS concepts towards different developer communities.

## 1 Introduction

Data science (DS) is a broad term that encompasses statistics, algorithms, machine learning, artificial intelligence, deep learning, and others. With all the different terminology present in today's research landscape, it can sometimes be confusing what exactly data science encompasses. By analysing the discussions of developers online, it is possible to track the popularity of topics, algorithms, and frameworks among developers to better understand the evolving landscape of DS.

Among the sites on the StackExchange network, StackOverflow [22] is the largest general Q&A site for developers. While StackOverflow is a generic Q&A site that contains posts in many different domains, Kaggle [11] is a website focused on data analysis, and also happens to contain a forum for developer discussion. In general, however, StackOverflow is more software engineering and application oriented, but Kaggle is almost exclusively for data analysts. This provides an interesting point of comparison to determine how DS topics vary between these two communities.

This analysis uses Latent Dirichlet Allocation (LDA) [6] combined with qualitative analyses to investigate DS discussion on both StackOverflow and Kaggle. A large-scale empirical study on 197836 posts from StackOverflow and Kaggle are analysed to better understand DS-related discussion across online websites. The key contributions are as follows::

1. This is the first research paper to investigate the relationship between DS discussion on a Q&A website like StackOverflow with competition-oriented community like Kaggle.
2. Twenty-four key DS topics are identified.
3. Source code to perform the analyses in this paper [9]. Full reproduction package (including data) also available [8].

**Paper structure:** Section 2 introduces the motivation behind this work. Section 3 describes StackOverflow, Kaggle and Latent Dirichlet Allocation. Section 4 describes the three research questions along with the methods and data used in this work. Section 5 presents the results of each research question, including data analysis, findings, and discussion. Section 6 describes the threats to validity. Section 7 covers the related work. Section 8 and 9 conclude and suggest future directions, respectively.

## 2 Motivation

There has not been a previous large-scale study on the DS-related topic discussions of developers in more than one online website. This is a particularly interesting area to examine, as it can provide insights into how discussion in different online communities can vary despite being in the same domain. In addition, examining characteristics of topics over time can provide insights into whether problems involving DS have changed significantly throughout the years, or whether certain topics are popular among developers depending on the community. By gaining a better understanding of these trends and nature of these online forums, it is possible to provide insight to the software research community, educators, and online programming sites on what areas of DS require better and higher quality resources for improving developers' understanding, as well as their productivity.

## 3 Background

**Data Science Posts on StackOverflow**
StackOverflow [22] is an online Q&A forum that enables users to share knowledge regarding programming and software engineering tasks. A StackOverflow post usually contains a unique identification (id), title, question, a list of tags and zero or more answers along with community metrics like the number of views, likes (score) and favourites. The original poster can tag their post using an existing selection of tags and specify one answer as the best/accepted one. In addition, users can comment on an individual post (either question or answer). In this analysis, the title, question, and answers of a post are concatenated together to form a sample. Comments have typically been excluded from StackOverflow LDA analyses [1–4, 8, 23] as they usually contain off-topic discussion.

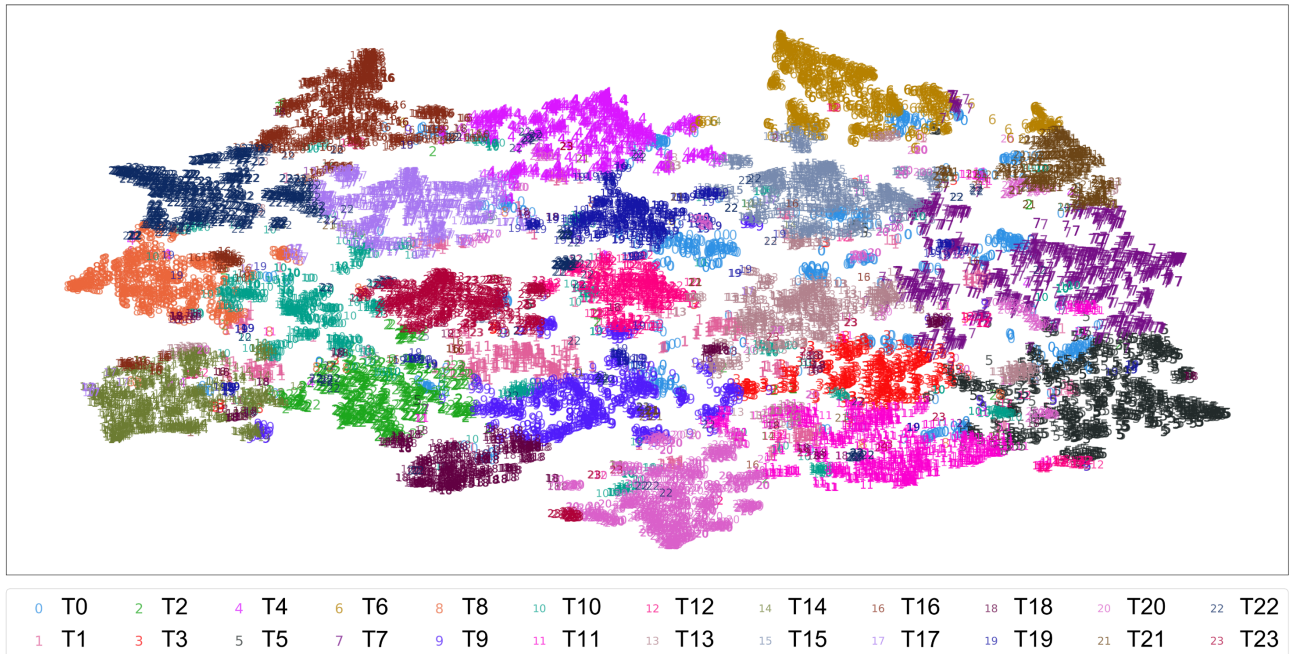

Figure 1: T-SNE performed on LDA topic probabilities for StackOverflow and Kaggle posts. Topic labels found in Table 2

A post is related to DS if it is broadly discussing any ideas relating to statistics, machine learning, data mining and other similar ideas. A much smaller emphasis was placed on purely theoretical mathematics and statistics questions, but it is likely that these posts are very rare on StackOverflow anyway since the focus is on programming, and there is a dedicated Mathematics StackExchange. The criteria (see Section 4.2) aimed to collect the largest sample of posts while minimising the false-positive rate.

**Discussion Threads on Kaggle**
Kaggle [11] is an online community that allows users to publish data sets, enter competitions to solve data science challenges, and provide a platform for data science discussions. In this study, only the "discussion" aspect of Kaggle will be analysed, as it is more comparable to a traditional Q&A website like StackOverflow. The Kaggle forums are split into six main categories: "generic", "getting started", "product feedback", "questions and answers", "datasets", and "learn". Posts from all these categories were included to provide a full picture of discussions on Kaggle. In the forums, users can create a thread or comment on a thread. Each thread has a unique id, and users can upvote comments or the original post. For this analysis, the title of a post, the post content and all its comments were concatenated to form a single sample. All threads on Kaggle were assumed to be related to DS.

**Latent Dirichlet Allocation**
The methodology used for uncovering the main discussion topics and their trends over time is known as Latent Dirichlet Allocation (LDA). LDA is a statistical topic model with the purpose of automatically identifying groups of related words that approximate a real-world concept (i.e. a "topic").

## 4 Case Study Setup

### 4.1 Research Questions and Methodology

There are three main Research Questions (RQs) to be answered to provide a broad picture on the current landscape of online discussion in StackOverflow and Kaggle in the DS domain. To answer these RQs, the method in section 4.2 is used to retrieve 197836 posts from both StackOverflow and Kaggle. This work was implemented in Python on a system running Intel Core i7-4200HQ CPU with 16GB of RAM.

**RQ1: What are the DS discussion topics on StackOverflow and Kaggle, and how do they differ?**
Currently, there have been no comparisons of the DS topics being discussed between StackOverflow and Kaggle. Following the practices of existing works [1–4, 8, 23], RQ1 uses topic modelling with Latent Dirichlet Allocation and manual validation to semi-automatically select the most interpretable set of DS discussions. Then, the proportion of the posts from each topic in each site is reported. These counts give insight into how the two forums differ regarding the most asked data science questions. Moreover, the second highest topic probability of a post will be used alongside the highest topic probability of a post to determine which topics are closely related, after filtering out all topic probabilities that are ≤0.15. This is described in Eq. (1).

$$TP(T_i, T_j) = \sum_{p \in D} \begin{cases} 1 & \begin{aligned} &if\ LDA(p, T_i) \Leftrightarrow \text{LDA}(p)_n \\ &\wedge LDA(p, T_j) \Leftrightarrow \text{LDA}(p)_{n-1} \\ &\wedge LDA(p)_{n-1} > 0.15 \end{aligned} \\ 0 & otherwise \end{cases} \quad (1)$$

Where TP(T_i,T_j) is the count for a pair of topics T_i and T_j, D are the DS posts from one site (StackOverflow or Kaggle), LDA is the trained LDA model, LDA(p,T_i) is the probability that post p belongs to topic i, and LDA(p) is the orderghb statistic of topic probabilities for post p.

**RQ2: How have the topics on StackOverflow and Kaggle evolved over time?**
RQ2 will report the number of posts in each topic by year for StackOverflow and Kaggle separately. Posts are included from the beginning of 2008 to the end of 2019. The year of a post is defined by the time the question (StackOverflow) or first post of a thread (Kaggle) was created; in other words, the creation time of subsequent posts for an answer or thread will not be accounted for. The counts for each year are non-cumulative, so that the rise or decline of topics can be more accurately represented.

**RQ3: How do the most mentioned ML/DL algorithms differ between StackOverflow and Kaggle?**
RQ3 identifies which algorithms are mentioned most, which can be used to determine the most generalisable or the most relevant algorithms. String matching (see Section 4.2) will be used to extract any mentions of a specific ML/DL algorithm from a pre-defined set of algorithms. The string matching will account for algorithms and their variants (i.e. abbreviations); for example, XGB and XGBoost will both be matched during string matching. In addition, if multiple string variants of the same algorithm appear in a post, they will be removed. The algorithm counts will be summed and grouped by their site of origin (StackOverflow or Kaggle), and then divided by the total sum of posts in either StackOverflow or Kaggle which contain at least one of the identified ML algorithms.

### 4.2 Data Science Post Collection

To study the characteristics of Data Science (DS)-related discussion in online forums, it is first necessary to obtain a reliable and large collection of posts related to DS. This is a trivial matter for Kaggle, as all posts from the meta-Kaggle dataset [16] are assumed to be DS-related. For both Kaggle and StackOverflow, a single sample is made of the concatenation of the post title, post question and post answers. For StackOverflow, the tags are also concatenated by space and included, but Kaggle does not have tags.

Data collection was not trivial for StackOverflow, since there can be posts that are not tagged correctly in StackOverflow. First, the data was downloaded using Google BigQuery. Existing works sometimes make use of a target tag [3] to identify a certain subset of tags, which are then used to filter related posts. In the case of DS, there is no specific DS tag on StackOverflow. In addition, there is the possibility of incorrect tagging. Hence, a heuristic string-matching approach was taken in combination with the tag-matching approach. This involved first creating list of relevant strings, and then performing the string matching on the StackOverflow dataset (~19 million posts).

The list of relevant strings was manually curated through a process of online research. The sources were Wikipedia machine learning page [17] and sklearn's [18] implemented algorithms that were deemed to be relevant (Appendix A).

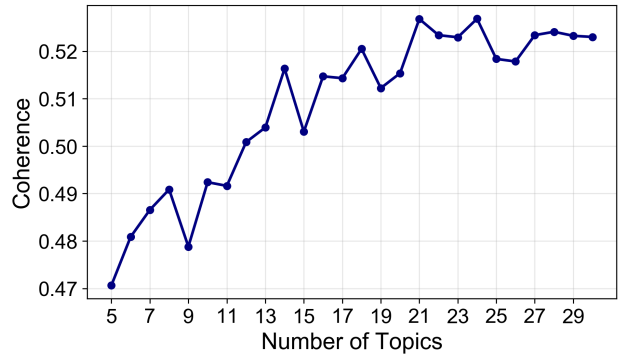

Figure 2: Coherence scores for LDA from grid search

Table 1: Hyperparameters used to tune Mallet LDA model. The best value for each parameter is bold

| Parameter | Range of values |
| --- | --- |
| Number of topics | 5, 6, 7, ..., **24**, ..., 29, 30 |
| Alpha | 0.01, 5, 10, **50** |
| Optimise Interval | 0, 20, **50** |

Due to size of the SO dataset, a fast, exact multi-pattern string searching algorithm was necessary. The chosen algorithm was the Aho–Corasick algorithm implemented in Python at [19]. Preliminary preprocessing was performed for this string search: this meant removing the HTML tags, all code blocks and snippets, removing certain special characters, stemming the words, and standardising UK/US spellings. This pre-processing was also applied to the Kaggle discussions. In addition, substring matching was enabled, but only for words of length larger than three (e.g. XGB, PCA). This is a custom-built string-matching package that is available at [10]. Then, the posts were filtered according to whether they had at least one of the matched words, resulting in approximately 500k out of 19 million posts. There was also a second stage of filtering to further refine this set—a post was included if it fulfilled either of two conditions: (1) the post contains at least three unique string-matched terms, or (2) the post contains a tag from a manually defined set of tags (Appendix B). This set of tags was determined through examining the common tags in the set of 500k, and through trial and error, determining the tags that had the largest number of posts. The tag list is short, since there were many data-science related tags that only contributed < 100 posts, and so would not significantly impact the final dataset.

Both StackOverflow and Kaggle posts were then pre-processed further. First, the language was determined using the Python package langdetect [12], to filter out all non-English samples, which are out of the scope of this analysis. Special characters were removed, and stop words were removed using the English stop word list from the NLTK Python library [5]. Then, the words were converted into bigrams and trigrams using the Gensim library with a min count of 5. This meant that words that appear together at least 5 times are joined with an underscore (e.g. "neural network" becomes "neural_network"). Finally, posts with less than 20 words or more than 1000 words were filtered out, as they are usually outlier posts that can reduce the quality of the topic models.

The final number of usable posts was 82065 posts from Kaggle and 115771 posts from StackOverflow.

**Table 2: List of topics related to DS on StackOverflow and Kaggle identified by Latent Dirichlet Allocation**

| Topic | Description | Key Words |
|---|---|---|
| T0: Generic errors | Generic troubleshooting and debugging in a machine-learning context. | code error run problem work python follow issue change line |
| T1: Training a model | Mainly related to questions about training, test and validation set. | datum set test training train model dataset data sample prediction |
| T2: Feature selection and regression | Questions involving either regression (e.g. linear, polynomial) and data type (continuous, categorical, numerical). | feature variable model datum column categorical regression target miss data |
| T3: Performance optimisation | Covers cost functions, algorithms like gradient-descent, regularisation, convergence, and reinforcement learning. | function loss gradient weight learn parameter learning rate update cost |
| T4: Dataset sharing online | Mainly regards meta discussion on Kaggle datasets, like where to find certain datasets, or a post providing a specific dataset(s). | kaggle kernel https dataset file notebook www download link github |
| T5: Input shape (tensors, arrays) | Questions about input type, usually to a deep learning model, like clarifications on manipulating input shape using TensorFlow. | shape input tensor tensorflow array lstm dimension output size batch |
| T6 TensorFlow installation troubleshooting | Usually, related to import errors, library version compatibility issues, and other exceptions that occur during TF installation. | tensorflow python file version error py install instal run line |
| T7: TensorFlow | A broad topic that encompasses many questions relating to TensorFlow. Includes troubleshooting and functionality clarifications | tensorflow model tf graph python variable code tensor create function |
| T8: Dataset-specific related discussion | Questions that are usually specific to the dataset. Some prevalent topics include dates/time, country, economics, and currency | datum user day year data product date month dataset predict |
| T9: Classification | All questions related to using classification (multi class, binary). | class classification classifier label machine svm scikit_learn learning model learn |
| T10: Processing scientific/field study data | Questions relating to a specific scientific context, like signal processing or experimental design. | time good make problem point question thing find work approach |
| T11: Neural Networks | Broad topic encompassing any neural-network related discussions. (e.g. includes feed forward, convolutional etc) | network neural layer output input learn weight deep neuron conv |
| T12: Natural Language Processing | Relates to the topic of NLP. Includes word embeddings, category predictions and document classification and preprocessing/filtering | word text document sentence vector nlp model embed feature question |
| T13 Keras | Broad topic relating to the Keras framework. | model keras kera train image loss training layer learn pytorch |
| T14: Kaggle leaderboards | Mainly Kaggle-focused discussions on leaderboard rankings. E.g. sharing scores and methods/models. | score lb model cv good feature public competition share fold |
| T15: Big data, memory, and GPU | Questions relating to processing big data. Includes technologies like Hadoop, Spark, Cloud Platforms, Kubernetes, map-reduce, clusters, GPUs. | gpu memory run time cpu machine process gb size datum |
| T16: Compliments | Primarily Kaggle threads that thank the original poster, e.g. if the visualisations or overall analysis is good. | kernel great work share nice good kaggle https www notebook |
| T17: Kaggle competitions, submissions, and teams | Primarily meta Kaggle discussions regarding competitions and submissions, and sometimes team building. | competition submission kaggle team submit score leaderboard make rule stage |
| T18: Evaluation and prediction | Questions about evaluation methods and metrics, often relating to prediction. Somewhat like T3, but more focused on evaluation. | score probability prediction metric distribution predict calculate model result give |
| T19: Data manipulation, dataframes, csv and files | Questions on how to work with dataframes, CSV data. E.g. merge, joining, reshaping, concatenation, reading data incorrectly | column file csv row datum data lt _ read table |
| T20: Computer vision | Questions relating to the field of computer vision. Can be about how to achieve a CV-related goal, improve detection performance, image segmentation, etc. | image opencv pixel object detect detection find face video feature |
| T21: OpenCV Java, Android, Caffe | A potential subset of T20, but with a large focus on Android/Java/Caffe technologies. | file opencv caffe java library build android project convert find |
| T22: Learning ML and DL | Threads where the focus is on learning. Can be generic, like how to start learning ML, or even some career advice threads. | learn datum machine learning data good https python science start |
| T23: Clustering | Threads and questions relating to clustering analyses. | cluster point distance datum matrix algorithm pca analysis number vector |

## 4.3 Topic Modelling with LDA

Because latent Dirichlet allocation (LDA) is an unsupervised algorithm, it is difficult to determine the correct number of latent topics. In addition, there are certain hyperparameters that can influence the performance of the LDA model. Hence, a grid search was utilised to find the optimum model. The Mallet implementation of LDA [14] was used and the parameters tuned are displayed in Table 1.

In total, 312 LDA models were trained. For each configuration, the coherence score was calculated, and the model with the highest coherence [20] for each topic number was plotted in Figure 1. Intuitively, coherence measures how coherent the content of posts are in the same topic. This resulted in two candidate models, which used either 21 topics or 24 topics. These two models were manually analysed to determine which topics were of higher quality. To do this, for each topic, the top-10 most frequent words were examined. In addition, a sample of the top-20 posts with highest probability for each topic were examined, alongside a separate random sample of 20. This procedure was also followed to label the topics. The best model was determined to be the model with 24 topics, which used an alpha value and optimise interval of 50.

To better visualise the separability of the topics, the output probabilities for the posts were plotted using the T-SNE clustering algorithm [13] (Figure 1) with the colours corresponding to the ground truth. topic of a post. To simplify analyses, the ground truth was defined as the topic of highest probability for a post predicted using the best topic model found. The separability of the clusters show that the topics are independent, and that the approximation of ground truth works well for most of the topics. Furthermore, this shows that the posts belong to a singular topic more often than being spread across many topics, which is consistent with the relatively high coherence score of 0.527.

## 5 Results

**RQ1**: *What are the DS discussion topics on StackOverflow and Kaggle, and how do they differ?*

Following the procedure in section 4.3, twenty-four DS topics were identified (Table 2). These topics were left completely unfiltered to best represent the natural state of discussions that occur in both sites. This meant that topics such as "Compliments" (T16) were not filtered out. However, a purely unfiltered approach also led to varying scopes for the topics. For example, "TensorFlow" (T7) arguably encompasses a larger variety of posts than T16.

Because LDA is an unsupervised approach, the clusters formed are expected to be distinct from each other. However, as seen in Figure 1, "T0: Generic Errors" does not form a distinct cluster. This is expected, since T0 is a generic topic that is likely to appear in many StackOverflow questions.

It can be also be seen that specific algorithms, frameworks and libraries were determined to be entire distinct topics ("T7: TensorFlow", "T11: Neural Networks", "T13: Keras"). This suggests that discussion based on these libraries and algorithms is potentially distinct from more generic machine learning topics, like "T9: Classification" and "T23: Clustering". In addition, two main areas of machine learning were identified

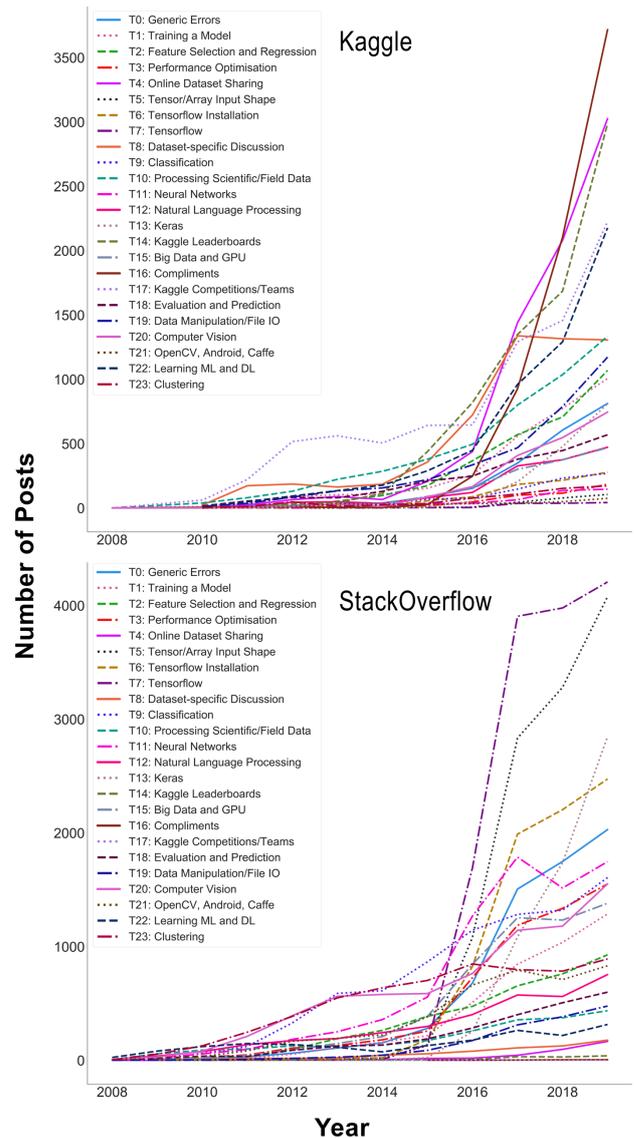

**Figure 3: Posts per topic from 2008 to 2019**

("T12: Natural Language Processing", "T20: Computer Vision").

The most common topic discussed in StackOverflow is T7 (14762 posts, 12.8%), while the most common topic in Kaggle is T16 (9668 posts, 11.8%) (Figure 4). In both cases, the samples are almost exclusive to one site – i.e. Kaggle has minimal discussion on TensorFlow, which could be due to the fact that TensorFlow discussions usually focus on lower level implementation details, which are not the aim of most Kaggle threads. This is further emphasised by the existence of the distinct "T6: TensorFlow installation troubleshooting", which suggest that the difficulty of setting up TensorFlow is significant. Interestingly, there is a higher proportion of T6 posts than T7 in Kaggle, which indicates that the TensorFlow discussions on Kaggle are primarily skewed towards set-up and installation.

There are a few other topics that are mainly discussed on only one of the sites. "T3: Performance Optimisation" is primarily discussed on StackOverflow; this is interesting, as it indicates that Kaggle threads are usually not distinctly focused on ideas like cost function, regularisation, and convergence. It is possible that this is due to the more practical nature of discussions on Kaggle, that are more focused on the dataset itself, as evident in the counts of posts for "T8: Dataset-specific

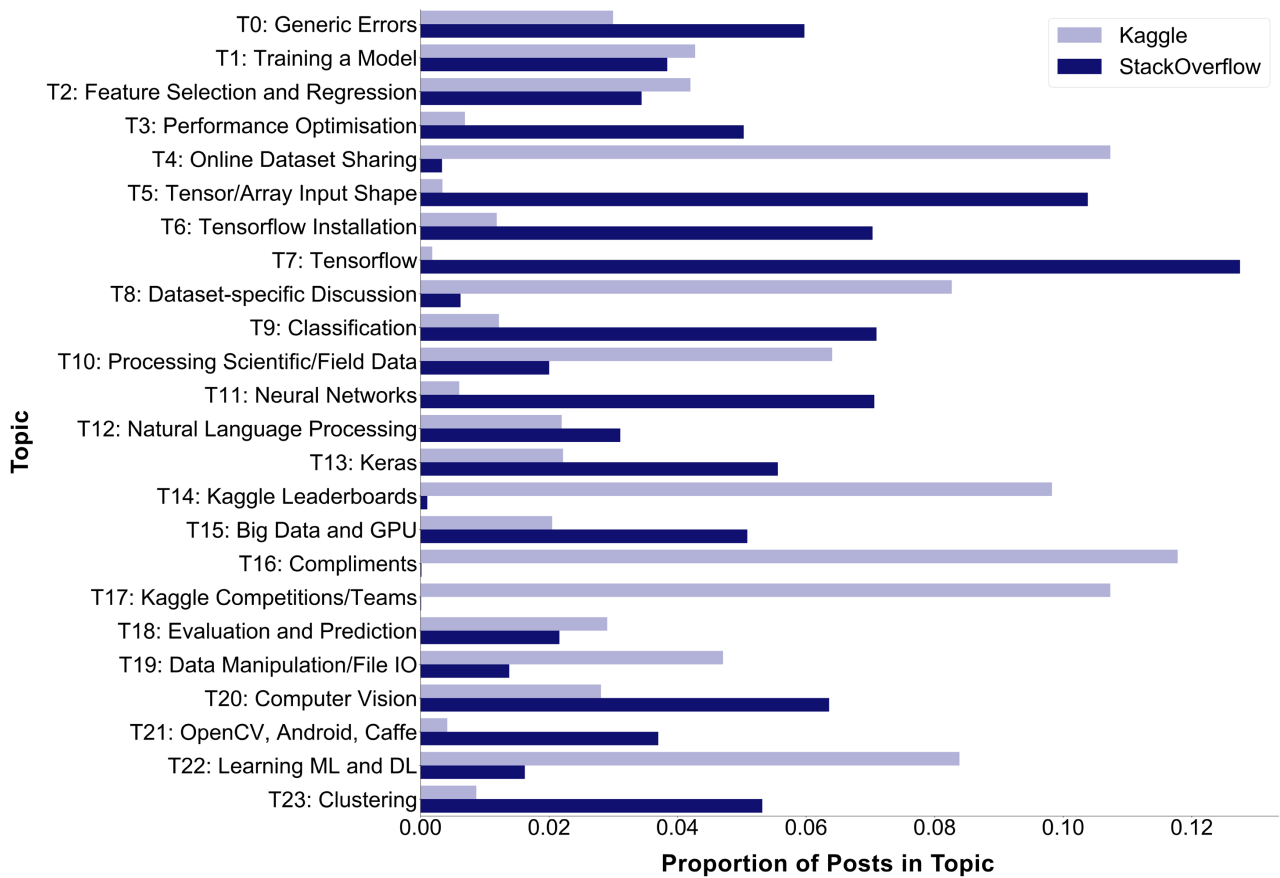

Figure 4: Proportion of posts per topic. Proportion is calculated as the sum of all posts in a topic divided by the sum of all posts in either StackOverflow or Kaggle.

discussion", which are mainly from Kaggle.

It should also be noted that the three primarily Kaggle-related topics T14, T16, and T17 are due to the nature of the site itself. The general purpose of StackOverflow is a Q&A website, but Kaggle is focused on competitions, which result in some unique "meta" discussion threads. T16 primarily consists of compliments based on visualisation, indicating that developers are favour data analyses workflows that include appropriate visualisations.

There are also numerous topics that are represented more equally on both sites. Of these, there are a few of interest. While T0 is primarily skewed towards StackOverflow, the counts in Kaggle are not insignificant, showing that troubleshooting threads are still present in Kaggle. "T1: Training a model" and "T2: Feature selection and regression" are both almost equally represented in both sites, which makes sense as they are more generic ML topics.

The most prevalent topic pairs in StackOverflow were T0 with T5, T6 and T7 (figure 5). This indicates that troubleshooting-based DS discussion on StackOverflow is primarily focused on TensorFlow. Furthermore, T7 occurs frequently with "T13: Keras" and "T15: Big data, memory and GPU", indicating that TensorFlow is closely related to working with big data. Logically, the relationship between Keras and TensorFlow is to be expected since Keras is a high-level API built on TensorFlow.

For Kaggle, the most prevalent pairs were T0 with T4, T10 with T8, and T22 with T16 and T17. This indicates that the general error discussions on Kaggle are more inclined towards dataset sharing, which suggests that the act of obtaining specific datasets could be viewed as a common problem in Kaggle. In addition, dataset-related discussion and processing scientific/field study data are closely related topics.

> **Answers to RQ1**: 24 DS topics were found in Kaggle and StackOverflow. TensorFlow-related topics were prevalent in SO, while meta discussion topics were the prevalent ones on Kaggle. StackOverflow tends to have more low-level troubleshooting discussions, while Kaggle discussions focus on practicality and optimising leaderboard performance.

**RQ2**: *How have the topics on StackOverflow and Kaggle evolved over time?*

Conforming to the results of RQ1, the most common overall topics in StackOverflow and Kaggle are also the most prevalent in recent times. In general, however, it can be seen that DS-related discussion in StackOverflow has risen dramatically in the past decade without signs of slowing down, indicating that the increasing popularity of DS is consistent with the amount of discussion that is occurring on StackOverflow. This is consistent with past research [3, 15]. Interestingly, the rise of TensorFlow-related discussion (both T6 and T7) appears to have slowed between 2017 and 2019, while discussion on tensors and input type have risen. This could be due to an oversaturation of TensorFlow-related questions, and/or adequate handling of duplicate posts. Furthermore, discussion on neural networks also appear to have peaked around 2017. However, the rise of the Keras topic has not slowed down but appears to be increasing towards 2019, suggesting shift in interest between these two frameworks.

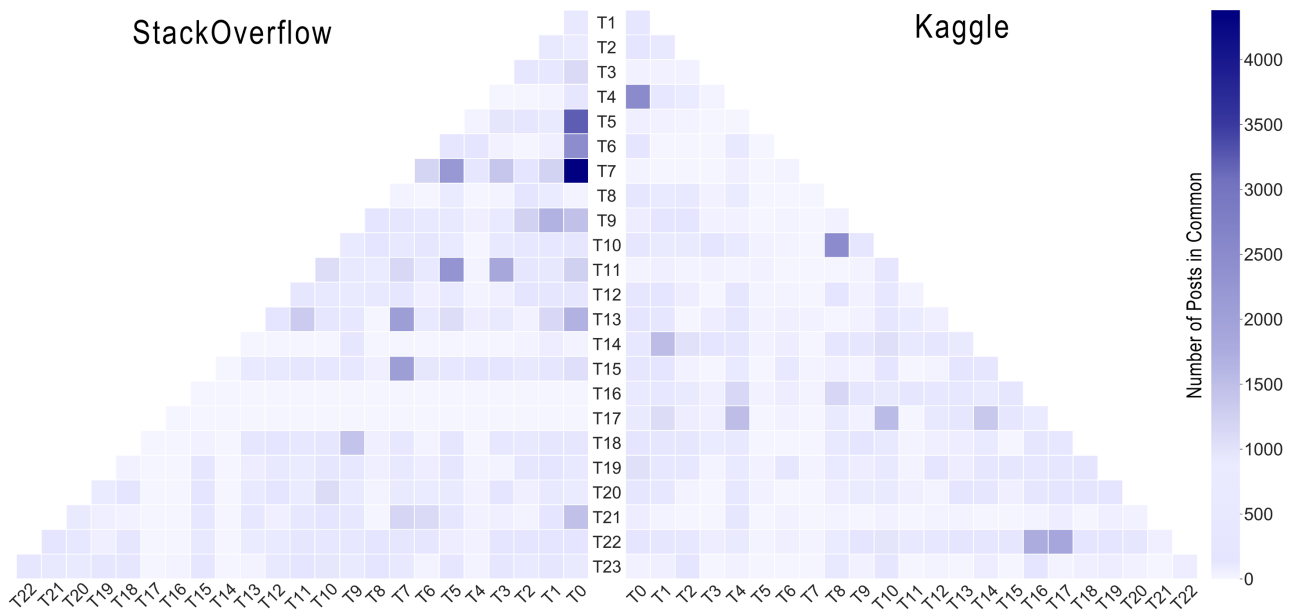

Figure 5: Shows the topic pairs in StackOverflow and Kaggle that are most represented. The colour scale is normalised using the largest topic pair count in both StackOverflow and Kaggle (T0/T7=4381).

There is a similarly dramatic rise in discussion on Kaggle, which indicates that the growth of the Kaggle site itself is only increasing. Unlike T7 in StackOverflow, T16 appears to be increasing at a dramatic rate. This suggests that the culture of Kaggle discussions is dominated by less technical discussion and is overall less regulated than the community of StackOverflow. In addition, while the meta-topic of competitions and teams was much higher in the earlier years of Kaggle, it has since been overtaken by three other topics. Most interestingly, dataset-related discussion appears to have slowed down in 2017. This could be due to a variety of reasons; one possibility is that the focus on the dataset itself is of lower priority than applying generalisable algorithms and visualisations in data analyses, or that datasets are better documented in recent times so that less discussion or questions are necessary.

> **Answers to RQ2**: DS discussion is increasing at a dramatic rate on both StackOverflow and Kaggle. TensorFlow discussion on StackOverflow appears to be slowing down, in comparison to Keras, which is rising rapidly. In Kaggle, the discussion topics rising most dramatically are non-technical and instead based on meta elements, like compliments.

**RQ3**: *How do the most mentioned ML/DL algorithms differ between StackOverflow and Kaggle?*

Fig. 6 highlights the differences in how often ML/DL algorithms are mentioned in either of the sites. The data is represented as a proportion, so that DS-related posts that do not contain any mention of a ML/DL algorithm are not included.

For StackOverflow, the most mentioned algorithm is Convolutional Neural Network (CNN), which is mentioned in 15% of all posts that mention a ML/DL algorithm. For Kaggle, the most mentioned algorithm is eXtreme Gradient Boosting, or XGBoost, which is mentioned in about 17.5% of all posts containing a ML/DL algorithm. There are numerous algorithms that are almost equally represented in both sites: these include linear and logistic regression, nearest neighbour algorithms, random forest, and principal component analysis. Some algorithms, like CNN, long short-term memory, Naïve Bayes, recurrent neural network, and support vector machine are more commonly mentioned in StackOverflow but are not insignificant in Kaggle. Light gradient boosting machine (LGBM) (10%) and XGBoost (17.6%) are commonly mentioned in Kaggle but are rarely ever mentioned in StackOverflow (both <2.5%). These are both ensemble learning methods, which usually are extremely generalisable to many supervised learning problems and result in high scores. These characteristics are extremely beneficial to the practical nature of Kaggle competitions, but the lack of discussion in StackOverflow suggests that developers do not have many questions to ask about these methods. Also, while AdaBoost is also an ensemble learning method, it is significantly less popular than LGBM and XGB in both StackOverflow and Kaggle.

> **Answers to RQ3**: Ensemble algorithms (LGBM, XGB) and CNN are the most mentioned ML/DL algorithms in Kaggle. CNN, LSTM and SVM are the most commonly mentioned ML/DL algorithms in StackOverflow. StackOverflow DS discussions rarely focus on ensemble methods.

## 6 Threats to Validity

The first threat relates to data collection in StackOverflow. The filtering method described in Section 4.2 relies on human judgement in multiple areas. First, manually curating a list of relevant strings is a subjective process. It is impossible to include every relevant term, since there is no formal definition regarding whether a term is DS-related. One attempt to reduce this threat was through the secondary filtering stage, which needed a post to either contain 3 or more unique DS terms, or to contain a tag that was specified to be DS-related. However, this secondary filtering heuristic was also determined through human judgement by attempting to find a balance between obtaining high-quality posts but also a reasonable number of posts to analyse. To further reduce this overall threat of data collection, having multiple people (preferably experts) check over this list would have been beneficial.

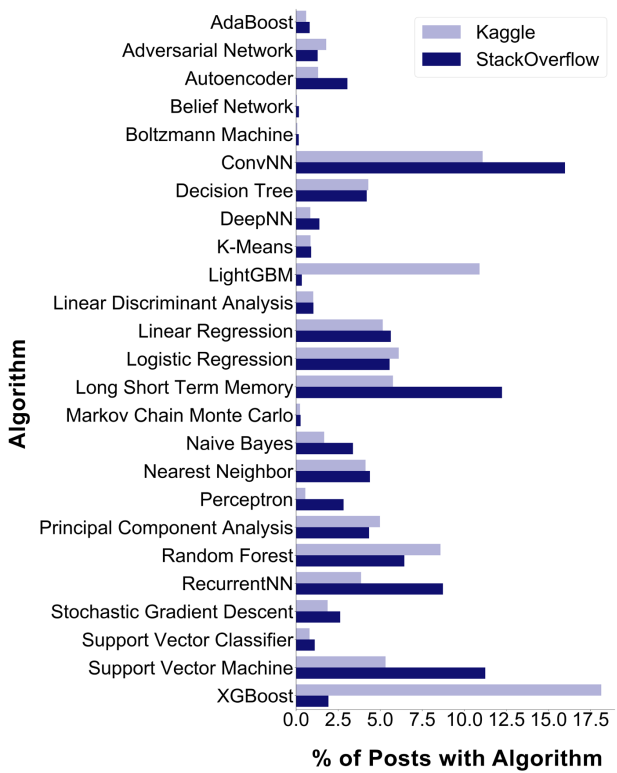

**Figure 6: String matching was used to determine the number of times a specific ML/DL algorithm appeared in a StackOverflow or Kaggle post. The percentage is out of the subset of StackOverflow or Kaggle posts that contain at least 1 ML/DL term.**

Language detection – the langdetect package works well but is not perfect. Through manual examination, it was found that certain Latin-based language comments in a thread were incorrectly identified as English. A better solution would be to use a more robust language checking algorithm, but this would come at the cost of time and efficiency. In addition, the number of posts missed by langdetect is likely insignificant, as the sites are primarily English-based, and the few non-English samples that were incorrectly identified did not appear to significantly influence the training of the LDA model.

In addition, since posts were only kept if their length were between 20 and 1000, it could introduce an artificial bias into the final posts that were analysed. In total, these outlier posts made up approximately 5k posts (~2.5% of all posts). Short posts were almost exclusively from Kaggle, and usually resulted from code dumps (which were filtered out, leaving only a few words), or short congratulatory / gratitude threads. For long posts, they mainly originated from StackOverflow, and were usually generic questions that drew in many answers without a clear topic. A separate analysis could be done on these large posts, but it is nonetheless a potential threat to validity in this study.

The 24 DS topics found may not be the best representation of topics. While the coherence was checked for number of topics from 5 to 30, Figure 2 suggests the possibility that coherence could have kept increasing past 30. While this may be the case, it would have also made the topics finer grained, which may not necessarily be more useful. For example, the TensorFlow topic could have been split into smaller topics, but it would not have added much to the overall conclusions. Potentially, further analyses could further focus on a topic like TensorFlow or Compliments by potentially performing a "nested" LDA analysis.

The labelling of the 24 DS topics found could also be of concern. Since this process was mainly subjective, the final labels were decided based on the interpretation of the author, who is only an undergraduate and not an expert. To minimise this threat, it would have been beneficial to have one to two other people, preferably experts in DS, to independently label the topics and provide input or discussion into how a topic should be labelled. Furthermore, the top 20 posts for a topic were examined alongside a random sample of 20; however, this is a small sample and could be increased further.

Finally, one other threat to validity could be the decision to include posts from all categories on Kaggle. While this gives the most realistic representation, it may not be the most comparable representation. The relevance of Kaggle posts could have been increased through additional filtering that only keeps technical-oriented discussions. This is a possible point that could be explored in future studies.

## 7 Related Work

Websites like StackOverflow contain many discussion posts in a wide variety of areas, and Kaggle provides a space where developers can discuss the best solutions to problems involving real-world datasets. Past research has primarily focused on StackOverflow due to its prevalence and usage among modern-day developers. Latent Dirichlet Allocation [6] has been commonly used to analyse the topics of different domains from StackOverflow.

Barua et al. [4] explored the topics of all posts on StackOverflow, and managed to identify better topics than the tagging system on StackOverflow. Many other studies have used LDA to analyse discussions in specific domain areas: security [23], mobile computing [21] and concurrency computing [1].

More relevantly, in 2019, Bangash et al. [3] conducted a study on ML discussions in StackOverflow using the Mallet implementation of LDA. However, to collect the machine learning posts, they only used posts with the "machine learning" tag, resulting in 28,010 posts, and did not focus on any other online discussion forums or websites. In addition, only two different number of topics (K=20, K=50) were tried.

Also in 2019, Bagherzadeh [2] focused on "big data" discussions from StackOverflow, and used a heuristic method [23] to obtain the relevant tag set which was related to big data discussions, obtaining 112,948 questions.

In 2020, Han [7] focused on three specific deep learning frameworks (TensorFlow, PyTorch and Theano), and examined discussions from StackOverflow and GitHub using LDA. They also examined the differences in the discussions between StackOverflow and GitHub regarding these frameworks.

These existing studies have not performed a large-scale and broad comparative investigation between DS topics discussed on Kaggle alongside StackOverflow. In addition, while there are some overlapping aspects with Bangash et al.'s [3] StackOverflow analysis, this work uses a much larger sample size. This resulted in the contradictory finding that TensorFlow is the most popular DS topic, as opposed to being the least popular topic when only considering the machine-learning tag [3]. This work also examines the frequency in which different ML/DL algorithms are mentioned, which is not explored in [3].

## 8 Conclusion

In this paper, 115771 posts from StackOverflow and 82065 posts from Kaggle were studied to highlight the difference in discussion of the same domain but in different communities. LDA was used to identify twenty-four commonly discussed DS topics on StackOverflow and Kaggle. Among these topics, it was discovered that TensorFlow-related topics were most prevalent in StackOverflow, while meta discussion topics were the prevalent ones on Kaggle. Moreover, while TensorFlow discussion on StackOverflow is slowing, interest in Keras is rising. In addition, the most mentioned ML/DL algorithms in StackOverflow are CNN, LSTM and SVM, while the most common algorithms in Kaggle are XGBoost and LGBM. Overall, StackOverflow discussions favour lower-level troubleshooting, while Kaggle discussions are aimed at practicality and optimising leaderboard performance. These findings will support educators and researchers with their understanding of DS topics among differing online communities.

## 9 Future Work

This study focused only on two popular online websites that contain developer discussion. However, there are many other forums where developers may discuss DS-related topics. The StackExchange network of sites itself contains the dedicated "Data Science" StackExchange and "Cross Validated" StackExchange, but these were not utilised in this study for the sake of scope. In addition, there may be DS-related discussion on sites like Reddit, Quora or even Twitter. In addition, there are other characteristics of topics that can be studied depending on the platform. For example, popularity and difficulty for StackOverflow posts can be determined through examining characteristics like "time to obtain accepted answer", "number of answers" and "reputation of answerers". This is not examined in this paper due to page limitations and scope. Future work could analyse online DS discussion for a much larger scope than only StackOverflow and Kaggle, and potentially more thoroughly investigate post and topic characteristics.

# Appendix A – String matching list

accordnet, active learning, adaboost, adadelta, almeidapineda recurrent backpropagation, alopex, analogical modeling, anomaly detection, anova, aode, apache singa, apriori algorithm, artificial intelligence, artificial neural network, association rule learning, association rules, autoencoder, automatic summarisation, averaged onedependence estimators, backpropagation, bayesian belief network, bayesian knowledge base, bayesian network, bayesian networks, bayesian statistics, biasvariance dilemma, binary classifier, bioinformatics, biomedical informatics, birch, boosted tree, boosting, bootstrap aggregating, bootstrap aggregating boosting, c45 algorithm, c50 algorithm, caffe, canonical correlation analysis, casebased reasoning, chisquared automatic interaction detection, classification, classification and regression tree, classifiers, cluster analysis, clustering, cn2 algorithm, cntk, collaborative and contentbased filtering, collaborative filtering, computer vision, conceptual clustering, conditional decision tree, conditional random field, constructing skill trees, contentbased filtering, convolutional autoencoder, convolutional neural network, cotraining, data analyses, data analysis, data mining, data preprocessing, data science, dbscan, decision stump, decision tree, deep autoencoder, deep belief networks, deep boltzmann machine, deep boltzmann machines, deep convolutional neural networks, deep learning, deep recurrent neural networks, deeplearning4j, dehaenechangeux model, denoising autoencoder, diffusion map, dimensionality reduction, discriminative model, dominancebased rough set approach, dynamic time warping, eclat algorithm, elastic net, email filtering, empirical risk minimization, ensemble averaging, ensembles of classifiers, errordriven learning, evolutionary multimodal optimization, expectationmaximization, expectationmaximization algorithm, extreme gradient boosting, extreme learning machine, facial recognition system, factor analysis, fastica, feature engineering, feature extraction, feature learning, feature selection, feedforward neural network, fishers linear discriminant, forwardbackward algorithm, fpgrowth algorithm, fuzzy clustering, gaussian naive bayes, gaussian process regression, gene expression programming, generative adversarial networks, generative learning model, generative model, generative models, generative topographic map, generec, genetic algorithm for rule set production, gensim, gradient boosted decision tree, gradient boosting machine, gradient descent, graphbased methods, graphical models, group method of data handling, growing selforganizing map, handwriting recognition, hidden markov model, hidden markov models, hierarchical classifier, hierarchical clustering, hierarchical hidden markov model, hierarchical temporal memory, hybrid recommender systems, hyper basis function network, id3 algorithm, idistance, image recognition, independent component analysis, inductive logic programming, information bottleneck method, information fuzzy networks, instancebased algorithm, instancebased learning, iterative dichotomiser, k means, keras, kernel methods for vector output, kernel principal component analysis, kmeans clustering, kmedians, knearest neighbor, knearest neighbors algorithm, knearest neighbors classification, knn, language recognition, lazy learning, leabra, learning automata, learning to rank, learning vector quantization, least absolute shrinkage and selection operator, leastangle regression, lgbm, lightgbm, lindebuzogray algorithm, linear classifier, linear discriminant analysis, linear regression, local outlier factor, logic learning machine, logistic model tree, logistic regression, logitboost, long shortterm memory, lowdensity separation, lstm, machine learning, machine translation, machine vision, manifold alignment, markov chain monte carlo, mcmc, meanshift, meta learning, minimum message length, minimum redundancy feature selection, mixture of experts, mlpack, model selection, multidimensional scaling, multilabel classification, multinomial logistic regression, multinomial naive bayes, multiple kernel learning, multivariate adaptive regression splines, naive bayes, naive bayes classifier, natural language processing, nearest neighbor, nearest neighbor algorithm, nearest neighbour, neural network, nonnegative matrix factorization, occam learning, online machine learning, optical character recognition, optics algorithm, ordinal classification, ordinary least squares regression, outofbag error, pac learning, partial least squares regression, pattern recognition, perceptron, prefrontal cortex basal ganglia working memory, preprocessing, principal component analysis, principal component regression, probabilistic classifier, probably approximately correct learning, projection pursuit, pvlv, pytorch, qlearning, quadratic classifiers, quadratic unconstrained binary optimization, querylevel feature, question answering, quickprop, radial basis function network, random forest, random forests, randomized weighted majority algorithm, recommendation system, recurrent neural network, regression, regression analysis, regularization algorithm, reinforcement learning, repeated incremental pruning to produce error reduction, ridge regression, ripple down rules, rprop, rulebased machine learning, sammon mapping, scikitlearn, search engine, search engine optimization, selforganizing map, semisupervised learning, singlelinkage clustering, skill chaining, sklearn, sliq, social engineering, sparse pca, speech recognition, speech synthesis, sprint, stacked autoencoders, stacked generalization, stateactionrewardstateaction, statistical learning, stepwise regression, stochastic gradient descent, structured knn, structured prediction, supervised learning, supper vector classifier, support vector machine, support vector machines, svc, svm, symbolic machine learning algorithm, tdistributed stochastic neighbor embedding, temporal difference learning, tensorflow, term frequency inverse document frequency, text mining, text simplification, tf idf, theano, torch, transduction, unsupervised learning, vc theory, vector quantization, wakesleep algorithm, weighted majority algorithm, xgb, xgboost

# Appendix B: StackOverflow Tags

deep-learning neural-network tensorflow keras pytorch kaggle machine-learning computer-vision caffe opencv pycaffe random-forest statistics dplyr k-means cluster-analysis pca bigdata